\documentclass[lettersize,journal]{IEEEtran}
\usepackage{amsmath,amsfonts}
\usepackage{array}
\usepackage{booktabs}
\usepackage{graphicx}
\usepackage{textcomp}
\usepackage{cite}
\usepackage{comment} 
\usepackage[colorlinks,urlcolor=blue,linkcolor=blue,citecolor=blue]{hyperref}

\DeclareUrlCommand\mdlraw{\urlstyle{tt}}
\DeclareRobustCommand{\mdl}{\mdlraw}

\newcommand{\tabnote}[1]{\par\vspace{3pt}%
  \parbox{\linewidth}{\footnotesize\raggedright #1}}

\begin{document}

\title{Can Prosodic Style Be Inferred from Text Alone? Evidence from
Unsupervised Acoustic Clusters}

\author{Abdul Rehman, Jian-Jun Zhang, and Xiaosong Yang%
\thanks{The authors are with the National Centre for Computer Animation,
Bournemouth University, U.K.}}

\markboth{Pre-print, Oct 2026}%
{Rehman \MakeLowercase{\textit{et al.}}: Prosody from Text}

\maketitle

\begin{abstract}
Much of expressive text-to-speech research rests on an untested
assumption that written text carries enough information to select an
appropriate prosodic style for its delivery. Text-predicted style models
improve listener preference, and expressive-appropriateness evaluation
presupposes that context constrains style, yet neither measures the
assumption itself. This paper tests it as a falsifiable hypothesis
against style labels derived from acoustics alone. For each of six
speakers in a 1{,}200-hour conversational corpus, utterances are
clustered in the spaces of five speech models, including a prosody-only
control, and the cluster of held-out utterances is predicted from twelve
text embedding models. Three controls are applied: utterance length is
erased from the speech embeddings;
accuracy is scored against the majority-class floor of unbalanced
clusters rather than uniform chance; and a bag-of-words baseline
measures word identity alone. Text predicts the cluster above that floor
for all six speakers ($+0.111$ top-3 accuracy), but bag-of-words
achieves three quarters of this. Sentence embeddings add only $+0.026$,
largest for encoders \emph{not} trained for sentence semantics and
reversed by tree-based probes for all others. Acoustic clusters are not
compact in text embedding space in any of 360 configurations. The
prosody-only space weakens the association for five speakers, but not
for the speaker showing it most strongly. Text thus informs these
delivery clusters mainly through word choice, whether as a cue to
prosody or as a marker of topic and recording situation, and
reference-free style selection cannot assume more.
\end{abstract}

\begin{IEEEkeywords}
Prosody, speech synthesis, sentence embeddings, style selection,
text-to-speech.
\end{IEEEkeywords}

\section{Introduction}
\label{sec:intro}

\IEEEPARstart{H}{uman} speech communicates through two coupled channels:
the linguistic content (\textit{what} is said) and its paralinguistic
realization (\textit{how} it is said). Modern text-to-speech (TTS)
systems handle the first channel well but rely on external conditioning
for the second: to obtain contextually appropriate prosody, many systems
require reference audio, explicit style labels, or manually selected
style tokens~\cite{wang2017tacotron, shen2018natural, wang2018style,
li2023styletts, lajszczak2024base}. A system asked to read an apology, a
sponsor announcement, and a casual aside should be able to infer from
the text that each calls for a different delivery, yet without explicit
style conditioning it typically cannot.

A substantial body of work suggests that this bottleneck is solvable
from the text side. A style representation predicted from text improves
listener preference over an unconditioned
baseline~\cite{stanton2018predicting}, and a growing family of
successors predicts style or prosody from progressively richer text-side
inputs: cross-sentence context~\cite{lei2023context,
guo24d_interspeech}, hierarchical emotion distributions deduced solely
from text~\cite{inoue2024hierarchical}, contrastively learned
text--prosody alignments~\cite{ye2023clapspeech}, and diffusion models
that sample prosody conditioned on text~\cite{li23j_interspeech,
oh2024diffprosody}. Work on expressive appropriateness likewise shows
that human judgments of whether a rendering is suitable depend on
discourse context rather than on the isolated
waveform~\cite{wang2026evaluating}. All of these results, however,
presuppose rather than measure the underlying premise: the prediction
targets are latent style embeddings or prosodic features co-trained with
the synthesis model itself, and the outcome measures are synthesis
quality and listener preference. None of this work quantifies how
strongly text constrains prosodic style against an explicit null,
separates a text--prosody association from the two trivial explanations
that any such measurement admits, utterance length and unbalanced label
priors, or measures how much of it is carried by word identity alone.
More broadly, much of TTS research implicitly expects style selection
from text to emerge given enough model scale and
data~\cite{kaplan2020scaling, sutton2024bitter}; that expectation has,
to our knowledge, never been tested directly. This paper provides that
test.

The question addressed is: \textbf{can prosodic style be inferred from
text alone?} Concretely, if the prosodic styles a speaker actually uses
are identified from acoustics without any text supervision, can those
styles be predicted from a text-only representation of what was said?
The defensible form of this hypothesis is probabilistic. A deterministic
mapping from text to prosody is arguably impossible: the same sentence
admits multiple contextually valid renderings depending on discourse
context and speaker intent, which are not recoverable from the text
itself~\cite{ladd2008intonational}. What is tractable is the weaker
claim that textual content is not prosodically \emph{neutral}, consistent
with long-standing evidence that intonation and speaking style reflect
discourse structure and communicative intent~\cite{gumperz1992,
cole2015}. The hypothesis under test and its null are:
\begin{itemize}
  \item[$H_1$:] Within an individual speaker's natural speech, textual
  content statistically constrains prosodic style: a text-only
  representation of an utterance predicts its acoustically derived
  prosodic style category better than utterance length and label
  frequency do, and those categories occupy distinguishable regions of
  text embedding space.
  \item[$H_0$:] Prosodic style, as identified by unsupervised clustering
  of acoustics, is independent of text beyond utterance length and label
  frequency: text representations do not beat those references, and the
  style categories are geometrically indistinguishable in text embedding
  space.
\end{itemize}

Word identity is treated as a reference level rather than as a confound.
Individual words can themselves cue delivery, as interrogatives,
intensifiers, and discourse markers do, so an association carried by
word identity would still be usable by a TTS system; but the same
association would also arise if a speaker's vocabulary merely tracks
topic and recording situation. A bag-of-words arm is therefore reported
alongside the sentence embeddings, and the question whether text is
informative at all is kept separate from the question whether
sentence-level representations add anything beyond the words they
contain.

\begin{figure}[!t]
\centering
  \includegraphics[width=0.55\linewidth]{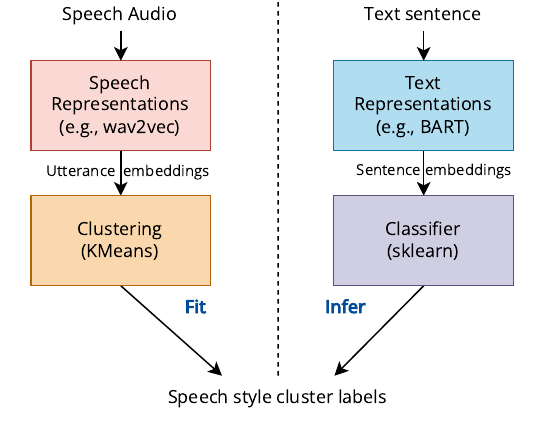}
  \caption{The cross-modal evaluation framework. Utterances are clustered
  using speech representation models (Table~\ref{tab:speech_models}); the
  cluster labels are then predicted from text embeddings of the
  transcriptions, and compared against length-only and bag-of-words arms
  that see the same labels and the same held-out utterances.}
  \label{fig:overview_text2style}
\end{figure}

The hypothesis is operationalized as a cross-modal cluster prediction
task (Fig.~\ref{fig:overview_text2style}). Each speaker's utterances are
embedded by a speech representation model and clustered; the cluster
index serves as an acoustically derived prosodic style label. Utterance
embeddings are mean-pooled over segments of up to 20~s, so these labels
describe utterance-level delivery, the register in which a stretch of
speech is produced, which is also the level at which reference encoders
and style tokens condition a TTS system~\cite{wang2018style}; local
intonation contours such as a final question rise are outside the scope
of the design. The
transcriptions of the same utterances are embedded by a sentence
transformer, and $H_1$ is confronted with three falsifiable predictions:
(P1)~text predicts the acoustic cluster of held-out utterances above the
majority-class floor and the length-only baseline, with sentence
embeddings adding to what bag-of-words achieves; (P2)~the acoustic
clusters occupy distinguishable, even if overlapping, regions of the
text embedding space; and (P3)~the alignment does not arise from lexical
content leaking through the speech representation, and therefore
persists when that representation contains no lexical channel. Each
prediction can fail independently. P3 addresses the most serious
confound of the design: general-purpose speech representations encode
phonetic, and hence lexical, content, so an apparent text--speech
alignment can be lexical leakage rather than prosodic coupling. The
speech-model pool therefore includes, alongside four self-supervised
models, a deliberately minimal prosody-only model (TIMBRESPACE,
Section~\ref{sec:timbrespace}) that discards phonetic detail by
construction.

The main contributions are:
\begin{enumerate}
  \item To our knowledge, the first test of the hypothesis that prosodic
  style is inferable from text against independently derived
  (unsupervised, acoustics-only) style labels rather than co-trained
  latents. The assumption exploited, implicitly or by construction, in
  prior work~\cite{stanton2018predicting, lei2023context,
  inoue2024hierarchical, ye2023clapspeech, li23j_interspeech,
  oh2024diffprosody, guo24d_interspeech, wang2026evaluating} is stated
  with an explicit null and evaluated across 5~speech representation
  models, 12~text embedding models, 6~classifier families, and 3~encoder
  depths, under a strict within-speaker protocol on six speakers. The
  speech encoders behind the labels are separately validated on labeled
  speech (Section~\ref{sec:iemocap}), where the same frozen
  representations decode four-way emotion at 0.53 to 0.59 weighted $F_1$
  against a 0.197 majority floor.
  \item A measurement protocol with the three controls such a test
  requires, and evidence that they are not optional: length-conditional
  erasure of the speech embeddings before clustering, which brings the
  length-only baseline down to the level of a constant predictor;
  scoring against the majority-class floor, which absorbs more than half
  of the apparent margin over uniform chance; and a bag-of-words
  baseline, which absorbs three quarters of what remains.
  \item The resulting bounded result. P1 survives in a weak form: text
  predicts its own utterance's acoustic cluster above the floor for all
  six speakers ($+0.111$ top-3), mostly through word identity, and
  sentence embeddings add only $+0.026$ over bag-of-words, positive for
  every speaker on average. P2 fails: acoustic clusters are not compact in text
  embedding space in any of 360 configurations.
  \item Two diagnostics on the character of the association. On the text
  side, \mdl{roberta-base} and \mdl{distilbert-base-uncased}, which
  are not trained for sentence similarity, give the largest increments
  and are the only encoders whose advantage over bag-of-words survives
  tree-based probes, while the 0.6B--1.5B instruction-tuned embedders
  rank last, which argues against a semantic reading of the increment.
  On the speech side, P3 holds only in part: the prosody-only
  representation retains less than half of the waveform encoders'
  association for five of six speakers, consistent with lexical
  leakage, yet most of it for the
  speaker with the strongest association, which lexical leakage cannot
  explain.
\end{enumerate}

\section{Background}
\label{sec:related}

\subsection{Self-Supervised Speech Representations}
Self-supervised (SSL) speech models learn representations from raw
waveforms that transfer well to supervised downstream
tasks~\cite{yang2021superb}. The three dominant families differ in their
pretraining targets: wav2vec~2.0 uses contrastive targets over masked
frames~\cite{baevski2020wav2vec}, HuBERT predicts iteratively refined
K-means cluster assignments~\cite{hsu2021hubert}, and WavLM adds speech
denoising and augmentation to masked prediction~\cite{chen2022wavlm}.
The ProsAudit benchmark~\cite{deseyssel23_interspeech} shows that SSL
models encode structural prosodic knowledge above chance, which supports
treating their embedding spaces as a source of prosodic structure.
Downstream benchmark success, however, does not imply that these models
organize speech into coherent \emph{unsupervised} structure: benchmark
rankings are sensitive to probing-architecture
choices~\cite{zaiem2024speech}, speaker and content information occupy
largely orthogonal subspaces~\cite{ashihara2024self,
qian2022contentvec}, and an analysis of ten pretrained SSL models by
Rehman \emph{et al.}~\cite{rehman2025boundaries} found that strong
classification performance frequently coexists with weak clustering
coherence for the same attribute. The present study responds with a
within-speaker clustering protocol, which controls for the dominant
speaker-identity structure in SSL embedding spaces, and with a
prosody-specific control representation alongside the general-purpose
waveform encoders.

\subsection{Prosody and Style in TTS}
Expressive TTS systems condition synthesis on style information obtained
from reference audio encodings~\cite{wang2017tacotron, shen2018natural,
wang2018style}, explicit pitch and duration
targets~\cite{ren2020fastspeech, lancucki2021fastpitch}, adversarial
style discriminators built on SSL
representations~\cite{li2023styletts}, large-scale speech-token language
modeling~\cite{lajszczak2024base}, prosody-focused self-supervised
objectives~\cite{hu2022prosodybert, liu2023speech_SER}, or discrete
latent style spaces~\cite{wang2019vector, kim2021conditional}. All of
these mechanisms answer \emph{how} to impose a chosen style; none
answers \emph{which} style should be chosen for a given input text.

A second line of work predicts the style from the text itself.
Text-predicted style models~\cite{stanton2018predicting} train a
predictor from text to the latent style embedding of a GST-Tacotron;
hierarchical context-aware predictors extend the text-side input to
neighboring sentences and previously realized
styles~\cite{lei2023context}; emotion-distribution predictors deduce
multi-level emotion intensities solely from
text~\cite{inoue2024hierarchical}; and diffusion-based predictors
replace deterministic regression with sampling, motivated by the
one-to-many character of the text-to-prosody
mapping~\cite{li23j_interspeech, oh2024diffprosody}. Closest of all are
contrastive frameworks that align text with prosody or style directly:
CLAPSpeech~\cite{ye2023clapspeech} contrasts the prosody of the same
spoken token across different text contexts, and
TACA~\cite{guo24d_interspeech} learns an utterance-level text-aware
style space under the supervision of a speech-derived one. Related
threads predict style embeddings from natural-language style
prompts~\cite{zhang2025histyle}, which supply the style description
explicitly, and model prosody from joint acoustic and textual input,
where the decoder is observed to rely chiefly on the textual
stream~\cite{eren25_interspeech}. The consistent success of this family
establishes that \emph{some} usable signal exists in text, but the
evidence is indirect: the prediction targets are latent variables
co-trained with the synthesis model or the alignment objective, the
outcome measure is synthesis quality or subjective preference, and,
with the partial exception of CLAPSpeech's same-token contrast, nothing
in these designs excludes lexical, length, or channel information as
the carrier of the learned association. From the evaluation side, Wang
\emph{et al.}~\cite{wang2026evaluating} formalize \emph{expressive
appropriateness}, whether given speech fits its discourse context, and
show that context-free judgments consistently underperform
context-aware ones; there, the text-to-style link is the premise of the
framework, not its object of measurement. None of this work measures
whether, and how strongly, text suffices to \emph{choose} the style,
which is the hypothesis test carried out here.

\subsection{Sentence Embeddings}
Sentence transformers~\cite{reimers2019sentencebert} map text to dense
vectors whose distances reflect semantic similarity, with variants
trained on natural language inference~\cite{lewis2020bart}, paraphrase
pairs, and general-purpose contrastive objectives~\cite{song2020mpnet,
wang2020minilm, nussbaum2024nomic}. Large language models and
retrieval-augmented generation~\cite{brown2020language,
lewis2020retrieval} would likely host an eventual deployment of
text-driven style selection. This study deliberately uses frozen text
embeddings with shallow classifiers instead, isolating the question of
whether the information is present and accessible from the engineering
question of how best to exploit it, and it includes plain
masked-language-model encoders alongside similarity-trained ones so that
``semantic'' and ``surface'' explanations of any effect can be told
apart.

\begin{table*}[!t]
\centering
\caption{Speech Representation Models}
\label{tab:speech_models}
\begin{tabular}{@{}lllll@{}}
\toprule
\textbf{Model} & \textbf{Params} & \textbf{Input} & \textbf{Pretraining data} & \textbf{Objective} \\
\midrule
HuBERT Large~\cite{hsu2021hubert} & 300M & waveform, 16 kHz & 60k h Libri-Light & masked prediction of K-means targets \\
HuBERT XLarge~\cite{hsu2021hubert} & 1B & waveform, 16 kHz & 60k h Libri-Light & masked prediction of K-means targets \\
WavLM Large~\cite{chen2022wavlm} & 300M & waveform, 16 kHz & 94k h mixed & masked prediction + denoising \\
wav2vec2 XLSR-53~\cite{conneau2021unsupervised} & 300M & waveform, 16 kHz & 56k h, 53 languages & contrastive \\
TIMBRESPACE (Section~\ref{sec:timbrespace}) & 52M & 6 prosodic features, 62.5 Hz & 860 h LibriSpeech & masked prediction of prosodic tokens \\
\bottomrule
\end{tabular}
\end{table*}

\section{A Minimal Prosody-Only Representation}
\label{sec:timbrespace}

Testing whether text--speech alignment is carried by prosody requires a
speech representation with ``just enough'' information about prosody and
timbre and as little as possible about lexical content. None of the SSL
baselines satisfies this: their waveform input retains full phonetic
detail. TIMBRESPACE is a simple transformer model built for this purpose;
code and checkpoints are publicly
available\footnote{\url{https://anonymous.4open.science/r/Timbrespace/}},
and further implementation details are given in the supplementary
material.

TIMBRESPACE operates on six frame-level acoustic features extracted at
62.5~frames per second (50~ms analysis frames, 16~ms hop): fundamental
frequency (F0), F0 bandwidth, first and second formant frequencies (F1,
F2), RMS power, and a formant-to-remainder signal-to-noise ratio (SNR),
each normalized to $[0,1]$ over fixed, perceptually motivated ranges.
The set covers the primary acoustic correlates of prosody and voice
quality while excluding phoneme-discriminative representations such as
MFCCs. F1 and F2 carry some vowel information, but the feature set as a
whole is drastically less lexically informative than spectrogram or
waveform input, which is what makes the model useful as a
lexical-leakage control. The frames are quantized into a vocabulary of
120 prosodic tokens by weighted K-means, and an 8-layer, 8-head,
768-dimensional transformer encoder is trained with BERT-style masked
language modeling over the token sequences, using span masking (1 to 10
tokens) and a curriculum masking probability increasing from 1\% to
20\%~\cite{bengio2009curriculum, hsu2021hubert}. Training uses
LibriSpeech~\cite{panayotov2015librispeech} in two phases
(train-clean-360, then train-other-500). Utterance-level embeddings are
obtained by mean pooling frame representations from a chosen layer,
identically to the SSL baselines.

Two properties matter here, both measured directly in
Section~\ref{sec:iemocap}. First, TIMBRESPACE embeddings cluster along
paralinguistic dimensions despite, or because of, their restricted
input. Second, speaker identity is the dominant axis in the space, as it
is in SSL embedding spaces~\cite{rehman2025boundaries} and in TTS
style-embedding spaces, which cluster first by speaker timbre and only
then by style attributes~\cite{zhang2025histyle}. Any cross-speaker
clustering would therefore produce speaker partitions rather than style
partitions, which motivates the strict within-speaker protocol used
throughout this paper.

\section{Methodology}
\label{sec:design}

\subsection{Task Definition}

For each speaker, let $\{(a_i, x_i)\}_{i=1}^{N}$ denote utterances with
audio $a_i$ and transcription $x_i$. A speech representation model $S$
produces utterance embeddings $S(a_i)$ (mean-pooled over the utterance's
real frames from a fixed layer), which are partitioned into $K=30$
clusters by K-means; the cluster index $y_i \in \{1,\dots,K\}$ is the
utterance's acoustically derived prosodic style label. A text embedding
model $T$ produces sentence embeddings $T(x_i)$. $H_1$ is tested by
(i)~training classifiers to predict $y_i$ from $T(x_i)$ on a train split
and evaluating on held-out utterances, against the same classifiers
trained on two reference feature sets (prediction P1), and
(ii)~measuring the geometry of the label partition $\{y_i\}$ directly in
the text embedding space (prediction P2). Clustering is fitted once per
(speaker, speech model, layer) and shared by all text models, so
text-model comparisons are free of clustering noise.

\subsection{Speech Representation Models}
\label{sec:speech_models}

Table~\ref{tab:speech_models} lists the five models. The four SSL
baselines span the three major self-supervised families and were the top
performers of their families in the ten-model analysis of Rehman
\emph{et al.}~\cite{rehman2025boundaries}; pretrained weights are taken
from the torchaudio pipelines. wav2vec2
XLSR-53~\cite{conneau2021unsupervised} additionally provides a
multilingual reference, and TIMBRESPACE the prosody-only control. All
models are used frozen. Layers are indexed proportionally on a 1--7
scale, index $i$ denoting the layer at $i/7$ of each model's depth, so
that a given index is the same relative depth in every model and index~7
is the final layer. The information carried by SSL layers varies
strongly with depth~\cite{pasad2024self}, so the index is a
consequential choice. All main results use the final layer. Indices~1
and~3 were also run, and Section~\ref{sec:results_layers} reports the
full comparison across the three depths.

\subsection{Text Embedding Models}

Twelve text embedding models are used: similarity-trained sentence
transformers (\mdl{all-MiniLM-L6-v2}~\cite{wang2020minilm},
\mdl{all-mpnet-base-v2}~\cite{song2020mpnet},
\mdl{paraphrase-distilroberta-base-v2}~\cite{reimers2019sentencebert},
\mdl{paraphrase-multilingual-MiniLM-L12-v2}); general-purpose dense
retrievers (\mdl{nomic-embed-text-v1.5}~\cite{nussbaum2024nomic},
\mdl{bge-large-en-v1.5}, \mdl{e5-large-v2}); an NLI-tuned encoder
(\mdl{bart-large-mnli}~\cite{lewis2020bart}); large instruction-tuned
embedders (\mdl{gte-Qwen2-1.5B-instruct},
\mdl{Qwen3-Embedding-0.6B}); and two plain masked-language-model
(MLM) encoders that are \emph{not} trained for sentence similarity,
\mdl{roberta-base}~\cite{liu2019roberta} and
\mdl{distilbert-base-uncased}~\cite{sanh2019distilbert}. The last
pair functions as a weak-encoder control: if the measured effect is
semantic, models trained to represent sentence meaning should beat
models that merely pool token embeddings.

\subsection{Classifiers and Reference Baselines}
\label{sec:classifiers}

Six scikit-learn classifiers, Logistic Regression, Random Forest,
K-Nearest Neighbors, Ridge Classifier, Linear Discriminant Analysis
(LDA), and Extra Trees, represent complementary geometric assumptions
and guard against the known sensitivity of representation evaluation to
the probing architecture~\cite{zaiem2024speech}. Features are
standardized with training statistics and the training set is subsampled
to 10{,}000 utterances; test sets are never subsampled. All reported
numbers use the \emph{best-performing} classifier per configuration by
weighted $F_1$.

This selection gives each arm its most suitable probe: LDA wins 359 of
360 text-embedding cells, whereas the TF-IDF arm is won by tree
ensembles in 228 cells and the length arm by Random Forest in 264.
Because it is made on the held-out split, it is a maximum over six
correlated estimates, and the selection gap (best minus mean over the
six classifiers) of 0.030, 0.022, and 0.014 weighted $F_1$ respectively
gives the scale of the resulting optimism in each arm's absolute level.
Every arm comparison is therefore also repeated with a single classifier
held fixed across arms (Section~\ref{sec:results_tfidf}), which removes
the selection step from the comparison.

Two reference arms run through the identical classifier suite on the
same labels and the same held-out utterances:
\begin{itemize}
  \item \textbf{Length only}: three features, the word count, character
  count, and heuristic syllable count of the text.
  \item \textbf{TF-IDF}: bag-of-words vectors, up to 2{,}000 features
  with sublinear term frequency, over scikit-learn's default word
  tokens (punctuation and single-character words excluded). This arm
  captures ``this is the cluster where the speaker says
  \emph{concrete}'' without any sentence-level semantics or surface
  form.
\end{itemize}
The length arm is the floor that any text representation must clear.
The bag-of-words arm measures what word identity alone achieves;
exceeding it is not required for text to be informative about delivery.

\subsection{Confound Controls}
\label{sec:confound_controls}

Three properties of the pipeline could create text-predictable clusters
with no prosodic style involved. Each is controlled, and the controls
change the result materially.

\subsubsection{Utterance Length}
Mean-pooled speech embeddings encode input length almost perfectly: a
ridge predictor recovers log utterance length with $R^2 \geq 0.97$ for
every model tested. This code is structural, not prosodic. A predictor
trained on natural utterances also recovers the duration of arbitrary
mid-utterance crops, which contain no phrase boundary, declination arc,
or final lengthening, with a predicted-versus-true slope of 1.01 to
1.12; excluding edge frames or subtracting per-position profiles does
not remove it. Without intervention, K-means therefore partly returns
duration bins. Two measures address this: embeddings are L2-normalized
before clustering, and the length-conditional mean is removed, that is,
clustering operates on $X - \mathrm{E}[X \mid \log \ell]$, with the
conditional mean fitted as a cubic spline on the training split only and
applied unchanged to test data. Erasure cuts the length-only baseline's
lift by 40 to 70\%. A residual association with length remains
($\eta^2 = 0.19$ to $0.52$), so the length arm is retained as a
diagnostic; at the final layer that residual is concentrated in the two
HuBERT models (Section~\ref{sec:results_p1}).

\subsubsection{Sequence Length Beyond the Trained Range}
TIMBRESPACE was trained on sequences of up to 622 tokens
($\approx$10~s), while 43.5\% of analysis sequences are longer, and
whole-sequence inference on those shifted embeddings further than the
distance between two unrelated utterances (median cosine 0.757 for the
longest third). Long utterances are therefore
encoded in overlapping 622-token windows with stride 311 and averaged
per token, which leaves short sequences bit-identical and brings the
model's length sensitivity down to WavLM's level. The waveform models
were verified to be stable over the same range (cosine $\geq 0.96$).

\subsubsection{Clustering Instability}
K-means at $K=30$ is sensitive to training-row order: shuffling rows
alone produced partitions with adjusted Rand index (ARI) 0.47 to 0.85
against each other. Runs therefore sort rows canonically, use float64
with ten initializations, seed all generators, and cache the fitted
clustering so that all twelve text models of a (speaker, speech model)
pair score the same partition. This makes every result reproducible,
and every comparison between text models or arms a comparison on one
fixed partition.

\section{Datasets}
\label{sec:datasets}

The analysis uses a corpus of Creative-Commons-licensed YouTube speech:
approximately 1{,}200 hours from six speakers (three male, three
female), subsampled to 200 hours per speaker to equalize speaker
effects, spanning diverse topics, registers, and spontaneous
conversational contexts. Per-channel attribution, hours per speaker, and
the data-use statement are given in the supplementary material.

Each source video is denoised before any further processing, with a
pretrained speech-enhancement model~\cite{defossez2020real} applied in
300~s windows with 10~s overlap and peak-normalized, and is then
transcribed with a Whisper-large-v3-turbo model using word-level
timestamp alignment~\cite{radford2023robust}. Denoising suppresses
background noise before any embedding is computed, which narrows the
recording-condition account discussed in Section~\ref{sec:discussion}.

Audio is segmented along transcript boundaries, merging adjacent
segments separated by at most 1~s up to a 20~s maximum, and discarding
segments shorter than a minimum duration, longer than the encoder's
maximum input, or carrying fewer than 5~word tokens per second, which
removes near-silent and sparsely transcribed spans; the transcriber's
own silence handling means segments do not contain long pauses. This
yields 22{,}929
to 66{,}619 training and 5{,}840 to 15{,}488 test segments per speaker.
The text attached to a segment is its own transcript; when that is
shorter than 200 characters, neighboring segments' text is prepended
and, if still short, appended, so that every text input carries at
least that much context.

Train--test partitioning is at the level of source videos, drawn
randomly but deterministically in an 80/20 ratio (426 to 1{,}270
training and 107 to 318 test videos per speaker), so no video
contributes utterances to both sides and classifiers cannot exploit
within-video shortcuts. Random rather than chronological selection also
removes any systematic era difference between training and test data; it
does not remove topic structure, since a channel's recurring settings
and subjects appear on both sides of the split.

All training and evaluation is performed on \emph{individual speakers}:
acoustic clusters are fitted independently on each speaker's data, and
text-to-cluster prediction is trained and evaluated within that same
speaker's utterances. The measured association therefore reflects how an
individual varies delivery with content, not differences between
speakers. The six speakers are accordingly treated as blocks, each a
complete replication of the experiment, rather than as a sample of a
population.

\section{Evaluation Metrics}
\label{sec:method}

\subsection{Classification Metrics and the Chance Level}

Classifiers trained on text features predict acoustic cluster
assignments on held-out utterances, scored by \textbf{top-3 accuracy}
(the proportion of utterances whose true cluster is among the three
highest-scoring predictions; the relevant metric for style selection,
where proposing a small candidate set suffices) and \textbf{weighted
$F_1$}, reported as a robustness check.

With $K=30$ clusters, uniform chance for top-3 is $3/K = 0.10$, but
K-means on speech embeddings does not produce balanced clusters (balance
scores 0.10 to 0.57 here, and near zero for one degenerate encoder). The
appropriate floor is therefore the \textbf{majority-top-3 rate}: the combined
share of a speaker's three largest clusters, which is the accuracy of a
constant predictor that ignores the text entirely. This floor averages
0.255 across configurations and reaches 0.448 for the most unbalanced
one, so an apparently large margin over uniform chance can be worth
nothing. All claims below are stated against this floor.

\subsection{Geometric Coherence Metrics}

Independently of any classifier, the acoustic-cluster partition is
evaluated directly in the text embedding space by the cosine
\textbf{silhouette} of the acoustic clusters computed on text embeddings
(positive values indicate that acoustically defined clusters are also
semantically compact) and by the \textbf{separation ratio} of mean
inter-cluster to mean intra-cluster distance. Raw coherence values are
not comparable across text models, whose cosine similarity scales
differ, so only within-model comparisons are made.

\subsection{Statistical Testing}
\label{sec:stats}

The unit of replication is the speaker. Cells are averaged within a
speaker first, so that speakers contribute equally, and one-sided
Wilcoxon signed-rank tests are then applied across the six speakers;
with $n=6$ the smallest attainable one-sided $p$ is $.016$ (all six
speakers agreeing) and five of six gives $.031$. Factor effects (choice
of text or speech embedding model) use Friedman tests with speakers as
blocks and Kendall's $W$ as effect size; layer comparisons use the same
cells at each depth, each scored against its own floor, with Friedman
tests over depths and two-sided Wilcoxon tests between them. Per-cell
binomial tests against the floor, and per-cell two-proportion tests
between arms, are also reported, but they treat held-out utterances as
independent Bernoulli trials; since consecutive utterances from the same
video are correlated, these $p$-values are \emph{nominal} and are read
as descriptive indicators of effect strength. Alongside the mean over
configurations, the \textbf{worst configuration per speaker} is
reported: the mean describes the typical configuration, and the worst
cell bounds it from below.

\section{Validation on Labeled Speech}
\label{sec:iemocap}


The style labels this paper predicts are unsupervised by design. To
verify that the embeddings behind them encode paralinguistic
information that listeners label consistently, we repeat the encoding
and clustering stages on
IEMOCAP~\cite{busso2008iemocap}, which supplies per-utterance emotion
and speaker labels, using the identical pipeline: same encoders, same
pooling, same length erasure, same K-means.

We use the corpus-provided speaker-disjoint split (Sessions 1--4 for
training, Session~5 for test): 3{,}289 training and 1{,}176 test
segments over the four standard emotion classes, with seven distinct
speakers surviving filtering in training and two entirely unseen
speakers in test, so every probe below is speaker-independent.
Embeddings are taken at the first and the final encoder layer. Because
the test split contains one male and one female speaker, speaker
identity and gender induce the same binary partition of the test set; we
report it as a single \emph{identity} axis. Supervised probes are the
classifier set of Section~\ref{sec:classifiers}, with the majority-class
predictor as the floor.

\subsection{Emotion Is Decodable from All Four SSL Encoders}
\label{sec:iemocap_probe}

\begin{table}[!t]
\centering
\caption{Speaker-Independent Probe Performance on IEMOCAP}
\label{tab:iemocap_probe}
\begin{tabular}{@{}lrrrrr@{}}
\toprule
& & \multicolumn{2}{c}{\textbf{Emotion}} &
\multicolumn{2}{c}{\textbf{Identity}} \\
\cmidrule(lr){3-4}\cmidrule(lr){5-6}
\textbf{Encoder} & \textbf{dim} & \textbf{L1} & \textbf{L7} &
\textbf{L1} & \textbf{L7} \\
\midrule
WavLM-Large      & 1024 & 0.570 & \textbf{0.586} & 0.984 & 0.949 \\
wav2vec2-XLSR53  & 1024 & \textbf{0.580} & 0.578 & \textbf{0.986} & \textbf{0.966} \\
HuBERT-XLarge    & 1280 & 0.537 & 0.582 & 0.983 & 0.935 \\
HuBERT-Large     & 1024 & 0.545 & 0.526 & 0.982 & 0.941 \\
TIMBRESPACE      &  768 & 0.417 & 0.397 & 0.814 & 0.834 \\
\midrule
majority floor   & --- & 0.197 & 0.197 & 0.692 & 0.692 \\
\bottomrule
\end{tabular}
\tabnote{Best weighted $F_1$ over the classifier set on held-out
Session~5. L1 and L7 are the first and final encoder layer. The identity
axis is the two unseen test speakers, who differ in gender.}
\end{table}

The four SSL encoders reach 0.53 to 0.59 weighted $F_1$ on four-way
emotion (Table~\ref{tab:iemocap_probe}), against a majority-class floor
of 0.197 (0.367 accuracy; class counts 432/323/260/161), clearing that
floor by 16 to 22 accuracy points in every cell
($p \le 1.4\times10^{-27}$, binomial). The representations the pipeline
clusters are not clustering on nothing.

These values sit below the accuracies above 70\% commonly reported for
IEMOCAP~\cite{liu2023speech_SER}, as expected: published numbers are
typically obtained by fine-tuning the encoder, pooling across layers,
and retaining utterance duration, whereas ours come from a single frozen
layer with the \emph{length confound erased}
(Section~\ref{sec:confound_controls}), and duration is a genuine
paralinguistic cue in acted corpora, not only a confound. At the first
layer, where the text baselines were logged on this corpus, a
length-only baseline reaches 0.111 to 0.195 top-3 against a
majority-class floor of 0.216 to 0.283, that is, \emph{below} a constant
predictor for all five encoders, as on the main corpus at the same
depth. Table~\ref{tab:iemocap_probe} is therefore a conservative lower
bound on the style information available to the pipeline, not a
speech-emotion-recognition benchmark.

\subsection{Identity Dominates the Embedding Geometry}
\label{sec:iemocap_identity}

\begin{table}[!t]
\centering
\caption{Unsupervised Structure on IEMOCAP}
\label{tab:iemocap_eta}
\begin{tabular}{@{}lrrrrrr@{}}
\toprule
& \multicolumn{3}{c}{\textbf{Emotion} $\bar\eta^2$} &
\multicolumn{2}{c}{\textbf{Identity} $\bar\eta^2$} & \\
\cmidrule(lr){2-4}\cmidrule(lr){5-6}
\textbf{Encoder} & \textbf{L1} & \textbf{L7} & \textbf{$\times$null} &
\textbf{L1} & \textbf{L7} & \textbf{ARI} \\
\midrule
WavLM-Large      & 0.056 & 0.055 & 21 & \textbf{0.808} & 0.144 & 0.009 \\
wav2vec2-XLSR53  & 0.056 & 0.035 & 14 & 0.470 & 0.019 & 0.007 \\
HuBERT-Large     & 0.058 & 0.036 & 14 & 0.550 & 0.039 & 0.013 \\
HuBERT-XLarge    & 0.049 & 0.041 & 16 & 0.367 & 0.059 & 0.014 \\
TIMBRESPACE      & 0.086 & \textbf{0.166} & \textbf{69} & 0.169 &
\textbf{0.161} & \textbf{0.024} \\
\bottomrule
\end{tabular}
\tabnote{Mean $\eta^2$ over ten UMAP components with no labels fitted;
``$\times$null'' is the L7 emotion $\bar\eta^2$ as a multiple of its
expected value under $H_0$. ARI: adjusted Rand index between the K-means
partition ($K=30$) and the emotion label at the final layer.}
\end{table}

At the first layer the identity axis is an order of magnitude stronger
than the style axis in every SSL space (Table~\ref{tab:iemocap_eta}): a
\emph{single} unsupervised component separates the two held-out speakers
at $\eta^2 = 0.71$ to $0.95$, while no component exceeds $\eta^2 = 0.12$
for emotion, and K-means purity against identity reaches 0.81 to 0.91
with no supervision at all. This is the empirical basis for the strict
within-speaker protocol: pooling speakers before clustering yields
identity partitions, not style partitions. The clustering in
Table~\ref{tab:iemocap_eta} is deliberately cross-speaker for that
reason; it is the contrast case, not the method.

\subsection{The Non-Lexical Control Behaves as Required}
\label{sec:iemocap_timbrespace}

TIMBRESPACE is the weakest encoder under supervised probing, which is
expected: 768 dimensions derived from six acoustic streams with no
phonetic channel cannot compete with 1024- to 1280-dimensional waveform
encoders at recovering categorical labels. What matters for its role is
whether its space is \emph{organized} along paralinguistic axes, and
under unsupervised measurement it is the most strongly organized of the
five. At the final layer it leads every label-free metric on both axes:
mean $\eta^2$ at 69 times the null, three times the strongest SSL
encoder and 4.7 times the weakest, maximum $\eta^2$ of 0.248,
cluster-label ARI and adjusted mutual information, and cluster--emotion
$\chi^2 = 506$ ($p<10^{-40}$), while separating the two unseen speakers
at $F_1 = 0.83$ from acoustic features alone. Because the pipeline
consumes these embeddings without supervision, this is the property it
actually depends on.

\subsection{Layer Choice Changes Geometry More Than Decodability}
\label{sec:iemocap_layer}

Between the first and final layer, SSL probe performance is essentially
flat (within $0.05$ $F_1$), but the underlying geometry is not: identity
$\bar\eta^2$ falls by 6 to 25 times and the cluster--identity ARI by 4
to 20 times, as the top layer specializes for the masked-prediction
objective. TIMBRESPACE is stable or improving across layers, and its
emotion $\bar\eta^2$ doubles at the final layer. A supervised probe is
therefore a poor instrument for selecting a layer for a clustering
pipeline, and the unsupervised side does not settle the choice either:
emotion $\bar\eta^2$ is equal or slightly higher at the first layer for
the four SSL encoders, whereas at the final layer their geometry is far
less dominated by voice identity and the prosody-only control is at its
most emotion-aligned. One encoder is degenerate at the final
layer: ten principal components carry 91.8\% of XLSR-53's variance,
against 33 to 54\% for the others, so its tight clusters (silhouette
$+0.18$) sit in a nearly low-rank space. The same collapse recurs on the
main corpus. Because labeled speech does not decide between depths, the
choice for the main experiments is made on the main corpus itself
(Section~\ref{sec:results_layers}).

\section{Results}
\label{sec:results}


The experiment comprises 360 configurations, 6 speakers $\times$ 5
speech models $\times$ 12 text models, all complete. Every configuration
uses $K=30$, the final encoder layer, and length-erased clustering. At
the final layer one speech model, XLSR-53, yields a degenerate partition
(Section~\ref{sec:results_speechmodels}). It is kept in every mean,
since the majority floor scores its imbalance correctly, and results
without it are given wherever they change a conclusion.

\subsection{Does Text Predict the Acoustic Cluster at All?}
\label{sec:results_p1}

\begin{figure}[!t]
\centering
  \includegraphics[width=0.99\linewidth]{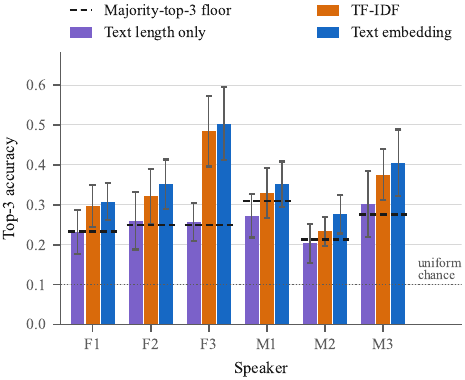}
  \caption{Top-3 accuracy of the three text arms per speaker, averaged
  over the twelve text models and five speech models (error bars:
  standard deviation over those cells). The dashed rule is each
  speaker's majority-top-3 floor; uniform chance (0.10) is shown for
  comparison only.}
  \label{fig:arms_by_speaker}
\end{figure}

All three arms exceed uniform chance in 360 of 360 cells, which
illustrates why that reference is uninformative here: so does a constant
predictor. Against the majority-top-3 floor
(Fig.~\ref{fig:arms_by_speaker}, Table~\ref{tab:perspeaker}), the text
embedding arm averages 0.366 top-3 against a floor of 0.255, a lift of
$+0.111$, and exceeds the floor in 293 of 360 cells (276 significantly
after Holm correction). At the speaker level it clears the floor for all
six speakers (Wilcoxon $p = .016$). The margin is smallest for M1
($+0.043$), whose clusters are the most unbalanced (floor 0.309). The
cells that fail are concentrated in one encoder: XLSR-53 clears its
floor significantly in 10 of 72 cells, the other four speech models in
50 to 72 of 72.

The length-only arm averages 0.254, which is \emph{at} the floor (mean
lift $-0.001$), exceeds it for three of six speakers, and is
significantly above it in 120 of 360 cells. Those cells are concentrated
in the two HuBERT encoders (60 and 48 of 72; mean lift $+0.035$ and
$+0.031$) and WavLM (12 of 72); for XLSR-53 and TIMBRESPACE the length
arm never significantly exceeds the floor. At the first layer the same
arm sat \emph{below} the floor for all six speakers
(Section~\ref{sec:results_layers}), so length-conditional erasure is
less complete at the final layer of the HuBERT models, where the
residual amounts to about a fifth of the embedding arm's lift. The
embedding arm nonetheless exceeds the length arm in 359 of 360 cells,
and for every speaker under each of the three linear probes.

\begin{table}[!t]
\centering
\caption{Per-Speaker Top-3 Accuracy of the Three Text Arms}
\label{tab:perspeaker}
\begin{tabular}{@{}lrrrrrr@{}}
\toprule
\textbf{Spk} & \textbf{Floor} & \textbf{Len} & \textbf{TF-IDF} &
\textbf{Sem} & \textbf{Sem--TFIDF} & \textbf{worst} \\
\midrule
F1 & 0.233 & 0.231 & 0.297 & 0.308 & $+0.011$ & $-0.058$ \\
F2 & 0.250 & 0.260 & 0.321 & 0.351 & $+0.030$ & $-0.037$ \\
F3 & 0.248 & 0.256 & 0.484 & 0.503 & $+0.019$ & $-0.038$ \\
M1 & 0.309 & 0.272 & 0.330 & 0.352 & $+0.022$ & $-0.040$ \\
M2 & 0.213 & 0.203 & 0.234 & 0.276 & $+0.042$ & $+0.004$ \\
M3 & 0.276 & 0.302 & 0.375 & 0.406 & $+0.030$ & $-0.020$ \\
\midrule
mean & 0.255 & 0.254 & 0.340 & 0.366 & $+0.026$ & \\
\bottomrule
\end{tabular}
\tabnote{Averaged over the twelve text models and five speech models.
Floor: majority-top-3 rate. Sem: text embedding arm. worst: minimum of
Sem--TFIDF over that speaker's configurations. Wilcoxon across the six
speakers: Sem $>$ floor, $p=.016$; Sem $>$ TF-IDF, $p=.016$; worst-case
Sem $>$ TF-IDF, $p=.98$.}
\end{table}

\subsection{What Carries the Association?}
\label{sec:results_tfidf}

The bag-of-words arm answers this directly. TF-IDF averages 0.340 top-3,
a lift of $+0.085$ over the floor, which is 77\% of the text embedding
arm's lift measured as a ratio of per-speaker means (median per-cell
share 86\%; 67\% excluding speaker F3). Sentence embeddings add $+0.026$
top-3 over TF-IDF on average ($+0.025$ weighted $F_1$), positive in 76\%
of cells (97\% on weighted $F_1$) and for all six speakers ($p = .016$
on both metrics). Per cell, the embedding arm beats TF-IDF significantly
in 187 of 360 configurations and loses significantly in 18.

\begin{table}[!t]
\centering
\caption{Increment of Text Embeddings over Bag-of-Words With the
Classifier Held Fixed}
\label{tab:fixedclf}
\setlength{\tabcolsep}{3.5pt}
\begin{tabular}{@{}lrccrr@{}}
\toprule
& \multicolumn{3}{c}{\textbf{All 12}} & \textbf{MLM} & \textbf{Sim.} \\
\cmidrule(lr){2-4}
\textbf{Fixed classifier} & \textbf{Sem--TFIDF} & \textbf{Spk $>0$} & $p$
& & \\
\midrule
K-Nearest Neighbors      & $+0.066$ & 6/6 & .016 & $+0.080$ & $+0.063$ \\
Linear Discriminant      & $+0.032$ & 6/6 & .016 & $+0.051$ & $+0.028$ \\
Logistic Regression      & $+0.020$ & 6/6 & .016 & $+0.037$ & $+0.017$ \\
Ridge Classifier         & $+0.019$ & 6/6 & .016 & $+0.036$ & $+0.015$ \\
Random Forest            & $-0.013$ & 0/6 & 1.00 & $+0.014$ & $-0.020$ \\
Extra Trees              & $-0.017$ & 0/6 & 1.00 & $+0.010$ & $-0.024$ \\
\midrule
best per configuration   & $+0.025$ & 6/6 & .016 & & \\
\bottomrule
\end{tabular}
\tabnote{Each row holds one classifier fixed across the embedding and
TF-IDF arms (weighted $F_1$, averaged within speaker; one-sided Wilcoxon
across the six speakers). MLM: \mdl{roberta-base} and
\mdl{distilbert-base-uncased}. Sim.: the seven similarity-trained
sentence encoders and the two instruction-tuned embedders. Last row:
the best-on-test selection used elsewhere in this paper.}
\end{table}

Two observations show how thin this increment is. First, it is not
uniform across configurations: in each speaker's \emph{worst}
configuration,
the embedding arm beats TF-IDF for only one of six speakers (M2,
$+0.004$; Wilcoxon $p = .98$), and on weighted $F_1$ for four of six.
Second, its sign depends on the probe and on the encoder
(Table~\ref{tab:fixedclf}). Under each linear probe the embedding arm
beats TF-IDF for every speaker. Under both tree ensembles, however, the
nine similarity-trained and instruction-tuned encoders lose to TF-IDF
for every speaker, because a tree can split on an individual informative
word, which is exactly the mechanism the bag-of-words arm was built to
capture. The nearest-neighbor row is not evidence either way, since it
is a poor probe for sparse high-dimensional TF-IDF. For those nine
encoders the defensible statement is therefore that their embeddings are
more \emph{linearly} separable with respect to the cluster labels than
bag-of-words vectors are, not that they carry more information about
them. The exception is the two MLM encoders, whose advantage survives
both tree ensembles ($+0.014$ and $+0.010$, for six and five of six
speakers): the only increment over bag-of-words that is robust to the
choice of probe comes from encoders that retain token identity and
surface form.

Across cells, embedding accuracy tracks bag-of-words accuracy almost
exactly (Spearman $\rho = 0.92$ over cells, $0.95$ over the 30
speaker--speech pairs): wherever text predicts the clusters, it does so
mostly through which words were used. Speaker F3 is the clearest case.
It has by far the highest predictability of any speaker (0.503 top-3),
but TF-IDF alone reaches 0.484, and on weighted $F_1$ the embedding arm
is only marginally above it (0.260 against 0.245). Its clusters track
vocabulary, an observation taken up in Section~\ref{sec:discussion}.

\subsection{Does Better Semantics Help?}
\label{sec:results_textmodels}

\begin{figure}[!t]
\centering
  \includegraphics[width=0.99\linewidth]{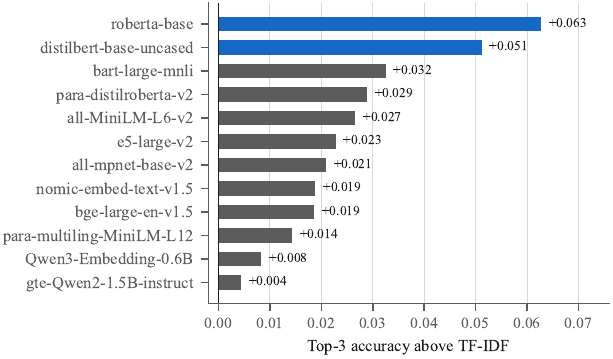}
  \caption{Top-3 accuracy above the TF-IDF baseline for each text
  embedding model, averaged over speakers and speech models. Highlighted:
  the two MLM encoders, included as weak-encoder controls.
  \mdl{para-multiling-MiniLM-L12} and \mdl{para-distilroberta-v2}
  abbreviate \mdl{paraphrase-multilingual-MiniLM-L12-v2} and
  \mdl{paraphrase-distilroberta-base-v2}.}
  \label{fig:text_models}
\end{figure}

If the residual beyond word identity were semantic, encoders trained to
represent sentence meaning should extract more of it than encoders that
merely pool token embeddings. The opposite holds
(Fig.~\ref{fig:text_models}). \mdl{roberta-base} ($+0.063$) and
\mdl{distilbert-base-uncased} ($+0.051$) lead every other encoder and
rank first and second for each of the six speakers individually;
\mdl{roberta-base} beats every non-MLM encoder in at least 29 of the
30 speaker--speech blocks. The seven similarity-trained sentence encoders
fall between $+0.014$ and $+0.029$ and the NLI-tuned
\mdl{bart-large-mnli} at $+0.032$, while the 0.6B and 1.5B
instruction-tuned embedders rank last ($+0.008$ and $+0.004$). The
text-model effect is itself significant (Friedman over the twelve
encoders, $\chi^2(11) = 48.5$, $p = 1.2\times10^{-6}$, $W = 0.74$ for
top-3; $\chi^2(11) = 48.9$, $p = 1.0\times10^{-6}$, $W = 0.74$ for
weighted $F_1$), so this ordering is not noise.

The natural reading is that the residual lives in \emph{surface form},
such as word order, function words, punctuation, disfluency, and
sub-word composition, which raw MLM token embeddings retain and
similarity-trained encoders deliberately normalize away; it shrinks
rather than grows with model scale, and Table~\ref{tab:fixedclf} shows
that only the MLM encoders keep their advantage under probes that can
split on single words. Part of that surface form may be prosodically
meaningful in its own right. The bag-of-words arm discards punctuation
(Section~\ref{sec:classifiers}), so a question mark that the MLM
encoders see is invisible to it; their advantage therefore argues
against a \emph{semantic} reading of the increment, not against a
prosodic one.

\subsection{Speech Models and the Prosody-Only Control}
\label{sec:results_speechmodels}

\begin{figure}[!t]
\centering
  \includegraphics[width=0.99\linewidth]{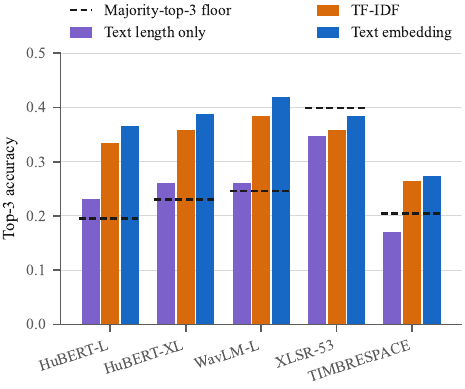}
  \caption{Top-3 accuracy of the three text arms per speech
  representation model, averaged over speakers and the twelve text
  models, against the corresponding majority-top-3 floor (dashed).}
  \label{fig:arms_by_speech}
\end{figure}

\begin{figure}[!t]
\centering
  \includegraphics[width=0.99\linewidth]{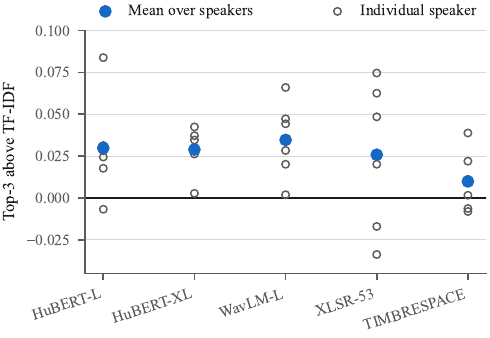}
  \caption{Top-3 accuracy above TF-IDF per speech representation model.
  Open circles are individual speakers (mean over the twelve text
  models); filled circles are the mean over the six speakers.}
  \label{fig:dsemtfi_by_speech}
\end{figure}

Prediction P3 states that the association should persist for a speech
representation with no lexical channel. It persists, but weakened
(Figs.~\ref{fig:arms_by_speech} and~\ref{fig:dsemtfi_by_speech}). The
three working SSL waveform encoders reach 0.365 to 0.419 top-3, a lift
of $+0.157$ to $+0.174$ over their floors, while TIMBRESPACE reaches
0.274, a lift of $+0.070$ that is significant in 50 of its 72
configurations but below each of the three waveform encoders for every
one of the six speakers (Friedman over the four encoders,
$\chi^2(3) = 11.6$, $p = .009$). The increment over TF-IDF is also
smallest for TIMBRESPACE ($+0.010$, against $+0.026$ to $+0.035$ for the
waveform encoders) and negative for two speakers, although at this depth
the increments do not differ significantly across speech models
(Friedman $p = .50$).

XLSR-53 is a separate case. Its final layer is nearly low-rank: ten
principal components carry 92\% of its variance, against 43 to 57\% for
the other encoders, and K-means returns a few large, tightly separated
clusters (acoustic silhouette $+0.29$, against $-0.08$ to $-0.05$
elsewhere; balance 0.01), the three largest holding 40\% of held-out
utterances. No text arm beats that floor on average (embedding-arm lift
$-0.015$), although the embedding arm still exceeds TF-IDF ($+0.026$).
The same collapse appears at the final layer on IEMOCAP
(Section~\ref{sec:iemocap_layer}) and is absent at the two earlier
depths (Section~\ref{sec:results_layers}), so it is a property of this
checkpoint's final layer rather than of the corpus or the pipeline. Its
well-separated but text-unpredictable clusters also show that cluster
compactness and text predictability are distinct properties.

The weakening is not uniform across speakers. For five of the six,
TIMBRESPACE retains between none and about a third of the three waveform
encoders' lift, the pattern expected if lexical content leaks from the
speech embeddings into the cluster labels. The sixth, F3, whose
association is the largest in the corpus, does not fit this account. The
prosody-only representation retains 81\% of the waveform encoders' lift
for F3 ($+0.273$ against $+0.338$), bag-of-words alone predicts its
clusters at $+0.262$ above the floor, and the ratio is 0.69 to 0.78 at
every depth of Section~\ref{sec:results_layers}. Structure that text
predicts in a representation without a phonetic channel cannot be
lexical leakage through the speech encoder. It must be carried by pitch,
formant, energy, or SNR differences: either by delivery, or by recording
conditions, which the two channel-sensitive features, power and SNR,
would register (Section~\ref{sec:discussion}). As a lexical-leakage
control, TIMBRESPACE therefore gives a speaker-dependent answer:
leakage can account for most of the association in five speakers, but
not in the speaker who contributes most of it.

The gap between TIMBRESPACE and the waveform encoders also co-varies
with model capacity. Across the 30 speaker--speech
pairs, the lift over the floor correlates with the width of the speech
embedding ($\rho = 0.38$ speaker-centered, $0.45$ without XLSR-53,
against $0.77$ at the first layer), and TIMBRESPACE, the narrowest
encoder (768 dimensions against 1024 and 1280), also produces the least
compact clusters, with roughly half the Calinski--Harabasz score of the
working SSL models. Wider encoders yield more separable clusters, which
are easier to predict from any input, so the weaker average association
of the prosody-only representation may in part reflect that narrower,
less separable clusters are harder to predict. Pretraining
domain does not explain the gap: the two HuBERT models, which show among
the strongest associations, were pretrained on read audiobook speech, as
TIMBRESPACE was. The compactness account is not the whole story either:
on labeled speech the same encoder produces the least compact clusters
and yet the \emph{best} alignment with emotion and speaker
(Section~\ref{sec:iemocap}), so low cluster quality does not by itself
imply a representation with little recoverable structure.

\subsection{Cluster Geometry}
\label{sec:results_geometry}

Prediction P2 is not supported. The cosine silhouette of the acoustic
clusters measured in text embedding space is negative in \emph{all} 360
configurations (range $-0.230$ to $-0.025$), with a mean separation
ratio of 0.253, above one in only 6 of 360: utterances sharing an
acoustic cluster are no closer in text space than utterances from
different clusters. This holds even though the coherence metrics are
computed on the most favorable subsample available, the at most 100
utterances nearest each cluster centroid. A classifier can still extract
a weak signal from systematic differences in mean location, but the
clusters are not ``about'' distinct content or registers in any
geometric sense.

The geometry is not uninformative, though. While the silhouette is, if
anything, inversely related to predictability ($\rho = -0.34$ across the
30 speaker--speech pairs), the \emph{separation ratio} tracks it
($\rho = 0.61$, and $0.58$ after centering within speakers). As a
reference level, the clusters are only weakly formed in the acoustic
space that defines them (mean silhouette $-0.08$ to $-0.05$ for the four
non-degenerate encoders), so compactness is a demanding criterion for a
30-way partition in either modality.

\subsection{Sensitivity to Encoder Depth}
\label{sec:results_layers}

\begin{table}[!t]
\centering
\caption{Sensitivity to Encoder Depth}
\label{tab:layers}
\begin{tabular}{@{}lrrrr@{}}
\toprule
\textbf{Layer index} & \textbf{1} & \textbf{3} & \textbf{7} & $p$ \\
\midrule
Majority-top-3 floor        & 0.219 & 0.220 & 0.255 & .011 \\
Embedding-arm top-3         & 0.343 & 0.361 & 0.389 & .030 \\
Lift over floor             & 0.124 & 0.141 & 0.134 & .042 \\
\midrule
\multicolumn{5}{@{}l}{\emph{excluding XLSR-53}} \\
\quad Floor                 & 0.217 & 0.221 & 0.219 & .61 \\
\quad Lift over floor       & 0.125 & 0.143 & 0.169 & .030 \\
\quad Weighted $F_1$        & 0.153 & 0.164 & 0.178 & .002 \\
\midrule
Speakers with lift $>0$     & 6/6   & 6/6   & 6/6   & \\
Speaker share of variance   & 0.87  & 0.75  & 0.52  & \\
Text-space separation ratio & 0.379 & 0.417 & 0.472 & .006 \\
\midrule
TF-IDF share of lift        & 0.71  & --    & 0.64  & \\
Length-arm lift             & $-0.022$ & -- & $-0.001$ & \\
Sem--TFIDF, weighted $F_1$  & $+0.024$ & -- & $+0.039$ & .016 \\
\bottomrule
\end{tabular}
\tabnote{The three text models run at every depth
(\mdl{roberta-base}, \mdl{distilbert-base-uncased},
\mdl{bart-large-mnli}): 90 matched configurations per depth, each
scored against the majority-top-3 floor of its own clustering. $p$:
Friedman test over the three depths with speakers as blocks; last row,
one-sided Wilcoxon of index 7 against index 1. The baselines were not
run at index 3.}
\end{table}

\begin{figure}[!t]
\centering
  \includegraphics[width=0.99\linewidth]{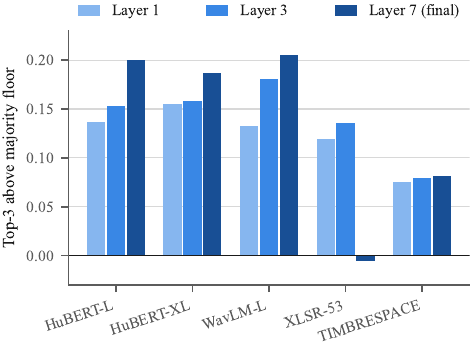}
  \caption{Top-3 accuracy above each configuration's majority floor, per
  speech representation model and encoder depth (index 1 is the first
  seventh of the model, 7 the final layer), averaged over the six
  speakers and the three text models of Table~\ref{tab:layers}.}
  \label{fig:layers}
\end{figure}

The final layer was fixed after indices 1 and 3 had also been run, so
all three are reported. Table~\ref{tab:layers} compares them over
matched configurations for the three text models run at every depth,
which are also the three strongest at the final layer, and
Fig.~\ref{fig:layers} breaks the lift down by speech model.

Excluding the collapsed XLSR-53, the final layer gives the largest lift
over the floor (Friedman $p = .030$) and the largest weighted $F_1$ for
all six speakers ($p = .031$ against either earlier depth), while the
floor itself does not move, so the gain is not an imbalance artifact.
Per encoder, the final layer gives the largest lift for five of six
speakers for each of HuBERT-L, HuBERT-XL, WavLM-L and TIMBRESPACE;
XLSR-53 peaks at index 3 and collapses at the final layer. The
association is also less concentrated in one speaker at depth: the
speaker share of variance in top-3 accuracy falls from 0.87 to 0.52, and
the weakest speaker's lift rises from $+0.016$ to $+0.074$.

The qualitative conclusions do not depend on depth. At every depth the
embedding arm clears the floor for all six speakers, and the acoustic
clusters have a negative silhouette in text space in every
configuration. Where the baselines were run, the embedding arm exceeds
TF-IDF for all six speakers at both depths, and \mdl{roberta-base}
and \mdl{distilbert-base-uncased} head the text-model ranking at
both. Two properties do change with depth: length erasure is more
complete at the first layer, and XLSR-53 is usable only below its final
layer. The final layer is used because it is where the association that
$H_1$ predicts is strongest for four of the five encoders, so negative
findings at that depth cannot be attributed to an unfavorable choice of
layer. P1, the positive result, is thereby reported at its most
favorable depth, and it holds at every depth examined.

\subsection{Verdict on the Null}
\label{sec:results_verdict}

$H_0$ is rejected, and what replaces it is a small, lexically carried
association. P1 is satisfied in a weak form: all
six speakers clear the majority floor ($p = .016$), three quarters of
that association is available from word identity alone, and the
embedding arm's advantage over bag-of-words, although positive for all
six speakers, holds for similarity-trained encoders only under linear
probes. The association varies more across speakers than across
models. Speaker identity
accounts for 55\% of the variance in top-3 accuracy across the grid,
against 25\% for the speech model and 3\% for the text model; one
speaker (F3) contributes a lift of $+0.255$ against $+0.082$ averaged
over the other five; and in each speaker's worst configuration the lift
over the floor is negative for all six, in every case an XLSR-53
configuration, and for two of six when that encoder is set aside. P2
fails in every configuration at every depth, and P3 holds only in part
(Section~\ref{sec:results_speechmodels}). The hypothesis that text alone
identifies prosodic style is therefore supported in a bounded form: text
is informative about these utterance-level delivery
clusters, mostly through which words are used, and sentence-level
meaning adds little.

\section{Discussion}
\label{sec:discussion}

\subsection{A Lexically Mediated Association}

The association that survives all three controls is real but narrow, and
its character is not what the TTS literature assumes: it is carried by
vocabulary. Two accounts of that vocabulary effect are consistent with
the results of Section~\ref{sec:results}, and they differ in what a TTS
system could do with it. In the first, a speaker's vocabulary
co-varies with the situation in which they are recorded: what someone
talks about changes with where, when, and how they are recorded, and the
acoustic embedding registers the accompanying differences in vocal
effort, distance, room, and microphone. The text side then predicts a
cluster without any prosodic mediation. Denoising eliminates the crudest
version of this channel story; microphone timbre, reverberation,
distance, and vocal effort survive it, and RMS power and SNR are two of
the prosody-only model's six features. The dominant speaker fits this
account: F3's
clusters are almost as text-predictable in a space of pitch, formant,
energy, and SNR features as in the waveform encoders.

In the second account, words are genuine cues to delivery.
Interrogatives, intensifiers, discourse markers, and punctuation signal
the register in which a stretch of speech is produced, so an association
carried by word identity would be prosodic coupling through the lexicon
rather than a confound, and would be usable by a TTS system as it
stands. The accounts make different predictions about which words carry
the association, content words, which track topic, or function words,
discourse markers, and punctuation, which cue delivery, and about how
strongly each speaker's clusters align with source video; both are the
natural next tests within this protocol. The results do not establish
that prosody is unpredictable from text. They establish that at the
granularity of this design, unsupervised 30-cluster partitions of
mean-pooled within-speaker utterance embeddings, and with trivial
explanations controlled, text informs delivery mostly through which
words are used.

\subsection{The Controls Are the Finding}
\label{sec:controls_matter}

Without the controls of Section~\ref{sec:confound_controls}, the same
data would have supported a strong positive result, and each control
removes a specific share of it. Pooled speech embeddings encode input
length structurally, so clustering them partly returns duration bins
that text length predicts trivially; after erasure the length arm falls
to the level of a constant predictor. The control is conservative:
duration is a genuine paralinguistic cue as well as a confound, so
erasing it biases every arm towards the null. The majority floor absorbs more than
half of the apparent margin over uniform chance: the collapsed final
layer of XLSR-53 reaches 0.384 top-3, nearly four times uniform chance,
without beating a constant predictor. The bag-of-words arm separates
``text is associated with delivery'' from ``sentence meaning is
associated with delivery'': without it, the 0.111 lift over the floor
would be attributed to semantics, when three quarters of it is word
identity. We recommend all three as minimum practice for claims of this
type, whatever their sign.

\subsection{Implications for Reference-Free TTS}

A retrieval mechanism that selects prosodic parameters by matching input
text against the transcriptions of each style cluster would, in this
corpus, operate at roughly 0.37 top-3 against a 0.26 constant-predictor
floor, with most of that margin carried by vocabulary. To the extent
that vocabulary reflects topic rather than delivery, such a mechanism
would retrieve the styles a speaker uses when discussing similar
subjects, which is not the same as the style the sentence calls for.
This bounds, rather than contradicts, the systems that exploit the
text-to-style link inside a synthesis
pipeline~\cite{ye2023clapspeech, guo24d_interspeech,
stanton2018predicting}: they co-train the style representation with the
objective, which may recover structure that unsupervised acoustic
clustering does not expose, and their gains are measured by listener
preference, which a weak but non-zero signal can still improve. What the
result does undercut is the expectation that scale and data alone will
deliver style selection from text~\cite{kaplan2020scaling,
sutton2024bitter}: the largest text encoders extracted the least beyond
bag-of-words.

\subsection{Limitations}
\label{sec:limitations}

The claims above are bounded by the design in the following ways.
\emph{Labels.} The style labels are unsupervised acoustic partitions,
not perceptually validated speaking styles. Section~\ref{sec:iemocap}
establishes that the embeddings being clustered carry decodable
paralinguistic information and that the prosody-only space is organized
along it, while categorical emotion accounts for little of a 30-way
partition (cluster--emotion ARI at most 0.024). What the clusters index
beyond that, delivery or recording context, is the question of
Section~\ref{sec:discussion}; per-cluster acoustic profiling and
listening tests would address it. \emph{Clustering.} Each (speaker,
speech model) pair uses one clustering at $K=30$, shared by all text
models and arms, so comparisons among them are made on identical labels,
whereas absolute levels would vary across seeds (ARI 0.47 to 0.85
between reorderings).
\emph{Probes.} The
classifier is selected on the held-out split, so absolute levels are
optimistic on the scale given in Section~\ref{sec:classifiers}, and the
central comparison is reported under fixed probes for that reason.
\emph{Speech models.} The speech-model comparison is not matched for
embedding width or cluster compactness
(Section~\ref{sec:results_speechmodels}). \emph{Corpus.} Each speaker is
one YouTube channel, which the within-speaker design holds fixed, but
recording situation varies within a channel and recurring topics and
settings appear on both sides of the split.
Transcriptions are ASR-generated, so every text arm follows the
transcriber's punctuation and disfluency conventions, ASR error patterns
co-vary with acoustic conditions, and the bag-of-words arm differs from
the embedding arms in access to punctuation as well as to sentence-level
representation. All data are English conversational YouTube speech from
six speakers, analyzed as blocks; read speech, other languages, and
population-level inference are outside the scope of this study.

These bounds notwithstanding, the test is made under the conditions of
practical use. The corpus is spontaneous, in-the-wild speech spanning
many topics, settings, and recording situations, which is the material a
deployed reference-free system has to handle, and a text-to-style link
robust enough to be relied upon should be measurable under such
conditions and not only in controlled read speech.

\section{Conclusion}
\label{sec:conclusion}

This paper turned a foundational assumption of expressive TTS, that an
appropriate prosodic style can be inferred from text alone, into a
falsifiable hypothesis with an explicit null, and tested it against
independently derived, acoustics-only style labels under three controls:
length erasure before clustering, a majority-class floor in place of
uniform chance, and a bag-of-words reference arm. Across 360
configurations spanning six speakers, five speech representation models
including a prosody-only control, twelve text embedding models, and six
classifier families, with three encoder depths compared, the result is a
weak, lexically mediated association. Text predicts its own utterance's
acoustic cluster above the majority floor for all six speakers ($+0.111$
top-3), but bag-of-words alone delivers about three quarters of that
lift. Sentence embeddings add $+0.026$, an increment that changes sign
with the probe for every similarity-trained encoder and is largest for
encoders not trained for sentence semantics. The association is weaker
in the prosody-only representation for five of six speakers, but not for
the speaker in whom it is strongest. Acoustic clusters are not compact
in text embedding space in any configuration, speaker identity accounts
for over half of the variance in the headline metric, and these
conclusions hold at each encoder depth at which they could be tested.

Within this design, then, text informs utterance-level delivery mainly
through which words are used. Whether those words act as cues to
delivery or as markers of topic and recording situation is the question
this result opens, and reference-free style selection cannot assume more
than this. The protocol extends directly to the analyses that would
answer it: separating content words from delivery cues in the
bag-of-words vocabulary, relating clusters to source video, and
perceptual validation of the clusters. Further extensions are listed in
the supplementary material.


\bibliographystyle{IEEEtran}
\bibliography{refs}

\end{document}